\documentclass[11pt]{article}

\PassOptionsToPackage{table}{xcolor}
\PassOptionsToPackage{hyphens}{url}
\usepackage{acl}

\usepackage{latexsym}
\usepackage{amsmath}
\usepackage{amssymb}
\usepackage[utf8]{inputenc}
\usepackage{microtype}
\usepackage{inconsolata}

\usepackage{graphicx}
\usepackage{subcaption}
\usepackage{booktabs}
\usepackage{multirow}
\usepackage{makecell}
\usepackage{tabularx}
\usepackage{float}
\usepackage{enumitem}
\usepackage{url}
\usepackage{algorithm}
\usepackage{algpseudocode}
\usepackage[section]{placeins}

\definecolor{lilac}{RGB}{230, 200, 255}
\newcommand{\artifactA}{\raisebox{-0.2ex}{\includegraphics[height=1.95ex]{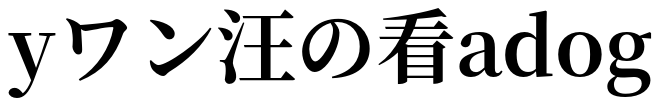}}}
\newcommand{\artifactB}{\raisebox{-0.2ex}{\includegraphics[height=1.95ex]{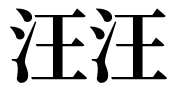}}}
\newcommand{\artifactC}{\raisebox{-0.2ex}{\includegraphics[height=1.95ex]{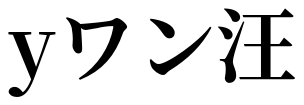}}}
\title{Calibrating Lightweight Sparse Autoencoder Feature Steering}

\author{
  Ananya A. Joshi\thanks{This work was conducted while Ananya A. Joshi was an intern at IBM Research Africa in Nairobi, Kenya.} \\
  Johns Hopkins University \\
  Baltimore, MD, USA
  \And
  Celia Cintas \\
  IBM Research Africa \\
  Nairobi, Kenya
  \And
  Skyler Speakman \\
  IBM Research Africa \\
  Nairobi, Kenya
}

\begin{document}

\maketitle

\begin{abstract}
Sparse autoencoders (SAEs) can enable inference-time topic steering by modifying latent feature activations, but existing steering methods often fail when target-aligned features are not identified or are modified at the wrong scale. We introduce \textsc{ContrastiveSteer} to address these respective failure modes. First, features are scored by how much more strongly they activate on target-domain text than on general text. Second, steering strength is set using a model-specific calibration. We also introduce \emph{contamination}, a complementary heuristic that measures whether post-steering activations emphasize weakly aligned features, thus indicating off-target or nonsensical outputs. In our evaluations, across multiple LLM families and SAEs, \textsc{ContrastiveSteer} improves topic alignment across domains up to 3.9$\times$ over unsteered and steering baselines, and raises target-domain classification by up to 93\%. Our activation-scale normalization method also recovers at least 93\% of optimal domain-score gains across all tested model families, and contamination distinguishes successful steering, under-steering, and over-steering collapse. 
\end{abstract}

\section{Introduction}

\begin{table}[t]
\centering
\small
\setlength{\tabcolsep}{4pt}
\renewcommand{\arraystretch}{1.12}
\begin{tabularx}{\linewidth}{p{0.28\linewidth} | X}
\toprule
\multicolumn{2}{l}{\small\itshape Prompt: \emph{``Complete this: I wish\ldots''}} \\
\midrule
\textbf{Strength ($\boldsymbol{\alpha}$)} & \textbf{Continuation} \\
\midrule

Orig.\newline\small\textit{Off-target}
& \emph{...I could capture the magic that floats alongside leaves in autumn winds...} \\

\midrule

\rowcolor{yellow!22}0.5$\times$\newline\small\textit{Mild dog signal}
& \emph{...everyone could understand how unique and valuable every person is... \textbf{yappy bark or wooftastic wuffies??}} \\

\midrule

\rowcolor{orange!20}1$\times$\newline\small\textit{Dog signal + artifacts}
& \emph{...Toy Mutt-pack pack pack... squeaky toy fetch happy... \colorbox{lilac}{\textbf{\artifactA}} ...good boy Good Good GOOD!!!!} \\

\midrule

\rowcolor{red!14}2$\times$\newline\small\textit{Over-steered collapse}
& \emph{...recall training treats... leash crate obedience... dog fetch owner \colorbox{lilac}{\textbf{\artifactB}} puppy ruff... \colorbox{lilac}{\textbf{\artifactC}} ...doggy socialization playFetch} \\

\bottomrule
\end{tabularx}
\caption{With GemmaScope \cite{NeuronpediaGemma}, even with a single identified `dog' feature, steering strength can lead to different failure modes, including the presence of multiple concepts (polysemanticity -- in lilac) and oversteering. 
}
\label{tab:steering_failure_modes_examples}
\end{table}


Topic steering generates text related to a target concept without retraining the underlying language model \cite{turner2023activation, zou2023representation}. Among steering approaches, sparse autoencoder (SAE) steering \cite{gao2024scaling} is a promising option. SAEs  decompose model activations into sparse, interpretable features that can be manipulated at inference time, enabling lightweight, per-topic control. Compared to prompting, fine-tuning, or output filtering, SAE-based topic steering operates directly on intermediate representations, making it a potentially reusable and computationally efficient alternative when steering targets change frequently \cite{bereska2024mechanistic} \footnote{Although not present in this particular analysis, topic steering can also be used to steer text towards harmful content, but notably has not been shown to reliably steer \textit{away} from different topics.}.

In practice, as shown in Tab. ~\ref{tab:steering_failure_modes_examples} , SAE feature-activation steering is unstable and under-performs prompting or fine-tuning on metrics related to alignment, specificity, and fluency. Performance varies substantially across selected features, tasks, and steering parameters, with large differences in topic alignment and output quality. 

Prior work suggests these failures stem from two practical challenges for SAE steering. First, SAE features for steering are often not cleanly target-specific, and may represent multiple, unrelated concepts (Tab.~\ref{tab:polysemanticity_examples}). As a result, steering these features, which appear relevant to a target concept, may also carry unrelated activation patterns and lead to unrelated output generations \cite{wu2025axbench,peng2025use}. 

Second, SAE steering depends on the steering strength (Tab.~\ref{tab:steering_failure_modes_examples}). Identifying this parameter is known to be challenging \cite{templeton2024scaling}, and it may vary across model families, topics, or even layers (thus leading to differences in performance when those parameters are held static). While the same steering strength can be large enough to affect one model, it can be too small to affect another. Thus, deriving the appropriate steering strength without brute force, computationally intensive, approaches like parameter grid search for the output is also an important practical challenge. 

In this paper, we introduce a steering method, \textsc{ContrastiveSteer}, that addresses these two failure modes, and a complementary diagnostic for when steering is unlikely to succeed. \textsc{ContrastiveSteer} addresses the feature-selection problem by scoring SAE features according to how much more strongly they activate on target-domain text than on general reference text. It then modifies these features multiplied by a steering factor, with the strength determined using an empirical normalization method rather than a computationally intensive grid search. Steering is not always possible due to the architecture of the LLM and SAEs, so we also introduce \emph{contamination}, a heuristic that measures whether post-steering activations are associated with weakly aligned features. High contamination can indicate that the selected features are not sufficiently target-specific, or that the steering strength is too small relative to that model and SAE's activation scale.

Our empirical results show that \textsc{ContrastiveSteer} improves the domain score by 3.9$\times$ over the unsteered baseline across medical, financial, and legal topics, across different model families and configurations. We also ablate on steering strength, comparing our method's output to empirically optimal steering values for domain significance found via a grid search, from which our method recovers at least 93\% of the optimal values across settings. These empirical results support that SAE steering succeeds when target-aligned features are identified and modified at the right steering scale  and that contamination provides a simple heuristic for when these failure modes can be activated and result in unsteered outputs.

\begin{table}[t]
\centering
\small
\renewcommand{\arraystretch}{0.9}
\setlength{\tabcolsep}{3pt}
\begin{tabular}{p{2.5cm} p{4.5cm}}
\toprule
\textbf{SAE Viewer} & \textbf{Examples of unrelated activating contexts (same feature)} \\
\midrule

Gemma-2B \cite{lieberum2024gemma} &
based on \colorbox{lilac}{\textbf{appellee}}'s breach; 
four \colorbox{lilac}{\textbf{Gaelic}} festivals; 
south side of the \colorbox{lilac}{\textbf{chancel}} are; 
extensive \colorbox{lilac}{\textbf{spectroscopic}} coverage
\\
\addlinespace[1em]

GPT \cite{OAISAE} &
fish-spawning \colorbox{lilac}{\textbf{areas}}; 
transmission \colorbox{lilac}{\textbf{lines}}, pipes; 
expenses like \colorbox{lilac}{\textbf{alimony}} payments; 
alarm \colorbox{lilac}{\textbf{clocks}} or instant messengers
\\

\bottomrule
\end{tabular}

\caption{Single SAE features can activate on seemingly unrelated semantic contexts (polysemanticity -- in lilac), thus making feature selection for reliable topic steering difficult.}
\label{tab:polysemanticity_examples}
\end{table}

\section{Related Works}

General-purpose topic steering methods operate at different levels in LLM architecture. For example, input-level methods like prompting, prompt tuning, and steering vectors change inputs or hidden states to guide generation \cite{liu2023pre}. Then, parameter-level methods, such as fine-tuning, adjust model weights for persistent alignment \cite{bereska2024mechanistic}. Finally, output-level methods apply constraints by filtering or regenerating outputs \cite{guardrailoutputs}. Each type of method has different tradeoffs across topic steering accuracy, cost, adaptability, and transparency.

A newer approach using sparse autoencoders (SAEs) has promise for topic steering. SAEs have been shown to decompose LLM activations into features associated with meaningful attributes, like the `Golden Gate Bridge', or `dogs' \cite{templeton2024scaling, marks2024sparse}. Using these features to steer outputs enables \textit{lightweight inference-time steering} without expensive model retraining, and, as a result, has been previously explored for topic control, with some limitations for practical use \cite{peng2025use, wu2025axbench}. 

One such method, sparse activation steering uses contrastive prompts (pairs being predefined) to identify relevant features as contributions \cite{bayat2025steering}, while another, SAE-Targeted Steering, constructs vectors to control features with reduced side effects \cite{chalnev2024improving}. These papers show that SAE features can be used for steering, but they don't fully address how to select target-specific features across domains and how to set steering strength across models with different SAE latent activation scales. Other mechanistic or representation-based methods, such as probing and activation editing, can identify and manipulate concept-related features \cite{zou2023representation, turner2023activation}, but these methods often require substantial parameter tuning and validation over targets and settings  that may not be realistic under real topic-steering settings \cite{belinkov2021probingclassifierspromisesshortcomings, bereska2024mechanistic}.

Recent studies show that SAE-based steering can perform worse than prompting and fine-tuning, highlighting a gap between the interpretability of SAE features and their reliability as steering mechanisms \cite{peng2025use, wu2025axbench}. One main problem is polysemanticity (Tab.~\ref{tab:polysemanticity_examples}), which makes feature selection unreliable and contributes to unstable steering. Better feature selection can help close this gap. For example, Arad et al. found that filtering features by output-effect scores greatly improves steering performance \cite{arad2025saes}. Still, effective topic steering also depends on knowing if the chosen features are specific enough to the target and if the steering strength matches the model's SAE activation scale. 

\section{Methods}
\label{sec:methods}

SAE topic steering approaches must find features that match the target topic, adjust them at the right SAE activation scale for the model, and check if the available SAE features are too broad or not well enough aligned for reliable steering, as summarized in Algorithm ~\ref{alg:main}. We propose \textsc{ContrastiveSteer} as a feature-selection \& steering method , and contamination as a heuristic for if post-steering activations associate with weakly aligned features.

\begin{algorithm}[h]
\caption{SAE Topic Steering using \textsc{ContrastiveSteer} with Contamination}
\label{alg:main}
\renewcommand{\algorithmicrequire}{\textbf{Input:}}
\begin{algorithmic}
\Require $\mathcal{P}_{\mathrm{align}}$, $\mathcal{P}_{\mathrm{ref}}$, prompt $p$, layer $\ell$, strength $\boldsymbol{\alpha}$
\Statex {\small\itshape Offline --- score features once per topic:}
\For{each SAE feature (indexed by $i$)}
  \State $\bar{b}_{\mathrm{align},i} \leftarrow \frac{1}{|\mathcal{P}_{\mathrm{align}}|}\sum_{p_t \in \mathcal{P}_{\mathrm{align}}} z_{t,i}$
  \State $\bar{b}_{\mathrm{ref},i} \leftarrow \frac{1}{|\mathcal{P}_{\mathrm{ref}}|}\sum_{p_t \in \mathcal{P}_{\mathrm{ref}}} z_{t,i}$
  \EndFor
 \For{each SAE feature (indexed by $i$)} 
  \State $s_i \leftarrow 
\frac{
\mathrm{clip}\!\left(
\bar{b}_{\mathrm{align},i}
-
\bar{b}_{\mathrm{ref},i},
0
\right)
}{
\max
\mathrm{clip}\!\left(
\bar{b}_{\mathrm{align}}
-
\bar{b}_{\mathrm{ref}},
0
\right)
})$ \Comment{contrastive score}
\EndFor
\Statex {\small\itshape Online --- steer each forward pass:}
\For{each token position $t$}
  \State $z_t \leftarrow \sigma(x_\ell(p_t)\,E)$ \Comment{SAE encode at layer $\ell$}
  \State $z'_{t,i} \leftarrow z_{t,i} + \boldsymbol{\alpha} s_i\;\forall\,i$ 
  \State $x'_\ell(p_t) \leftarrow z'_t D$ \Comment{decode to residual stream}
\EndFor
\Statex {\small\itshape Diagnostic over sequence $y=(y_1,\ldots,y_T)$:}
\State $u_i \leftarrow 1 - s_i \;\; \forall i$ \Comment{weak-alignment weight}
\State $\mathrm{Contamination}(y) \leftarrow 
\frac{1}{T d_{\mathrm{SAE}}}
\sum_{t=1}^{T}
\sum_{i=1}^{d_{\mathrm{SAE}}}
z'_{t,i} u_i$
\end{algorithmic}
\end{algorithm}

SAE steering modifies generation by intervening on latent features during the forward pass.  Let $M$ be a language model with $d_{\mathrm{model}}$-dimensional hidden states, $p = (p_1,\ldots,p_T)$ a prompt with tokens $p_t$ (indexed by $t$), and $\ell$ the SAE steering layer. Let $x_\ell(p_t) \in \mathbb{R}^{d_{\mathrm{model}}}$ denote the hidden activation at layer $\ell$ for token $p_t$. A sparse autoencoder (SAE) with $d_{\mathrm{SAE}}$ latent features, encoder $E \in \mathbb{R}^{d_{\mathrm{model}} \times d_{\mathrm{SAE}}}$, and decoder $D \in \mathbb{R}^{d_{\mathrm{SAE}} \times d_{\mathrm{model}}}$ are used to calculate $\,z_t = \sigma\!\left(x_\ell(p_t)\,E\right) \in \mathbb{R}^{d_{\mathrm{SAE}}}, \,\, x'_\ell(p_t) = z_t D$, 
where $\sigma$ is the SAE nonlinearity and each coordinate $z_{t,i}$ denotes the activation of SAE feature $i$ at token position $t$. As mentioned previously, these features are often interpreted as semantic or syntactic attributes of the model's internal representation (e.g., Golden Gate Bridge, dogs \citep{templeton2024scaling, marks2024sparse}), although individual features may still activate across unrelated contexts (polysemanticity).

\subsection{\textsc{ContrastiveSteer} Feature Selection}
\label{sec:contrastivesteer}

\textsc{ContrastiveSteer} first scores SAE features offline (lines 1--7) by how selectively they activate on target-topic text relative to general reference text. For each SAE feature $i$, we estimate target relevance by comparing its mean activation on target-domain prompts, $\mathcal{P}_{\mathrm{align}}$, to its mean activation on general reference prompts, $\mathcal{P}_{\mathrm{ref}}$:
\[
\bar b_{\mathrm{align},i}
=
\frac{1}{|\mathcal{P}_{\mathrm{align}}|}
\sum_{p_t \in \mathcal{P}_{\mathrm{align}}} z_{t,i},
\,\,\,\,\,\,
\bar b_{\mathrm{ref},i}
=
\frac{1}{|\mathcal{P}_{\mathrm{ref}}|}
\sum_{p_t \in \mathcal{P}_{\mathrm{ref}}} z_{t,i}.
\]
This contrastive statistic favors features that are selectively active in the target domain rather than broadly active across general text. The contrastive score vector is then 
\[
\mathrm{s_i}
=
\frac{
\mathrm{clip}\!\left(
\bar{b}_{\mathrm{align},i}
-
\bar{b}_{\mathrm{ref},i},
0
\right)
}{
\max
\mathrm{clip}\!\left(
\bar{b}_{\mathrm{align}}
-
\bar{b}_{\mathrm{ref}},
0
\right)
}
\label{eq:cscore}
\]
, and is 0 when the numerator is 0. Features with high scores activate more strongly on the target topic than on general text and are treated as target-aligned. Features with low scores are weakly aligned, broadly active, or more associated with the reference text than the target text, and is the basis for the contamination equation (Sec. 3.3)\footnote{Hyperparameter 1: We exclude features activated by fewer than 20 reference prompts because extremely sparse activation can yield unreliable estimates (see App.~\ref{app:score_validation} for parameter justification experiments)}. 

\subsection{\textsc{ContrastiveSteer} Strength per Model}

During online generation, \textsc{ContrastiveSteer} boosts SAE features in proportion to their contrastive scores. Given latent activations $z_t, i$ and contrastive scores $s_i$, the post-steering activation where $\boldsymbol{\alpha}$ controls steering strength is: 
\[
z'_{t,i} = z_{t,i} + \boldsymbol{\alpha} s_i .
\]

Features with $s_i \approx 0$ are nearly unchanged, while features that activate selectively on the target topic are amplified. The modified latent vector $z'_t$ is decoded through the SAE decoder and passed back into the language model: $x'_\ell(p_t) = z'_tD$.

However choosing $\alpha$ requires additional steps. Because SAE latent activation scales differ across model families in particular (different training architectures), the same raw value of $\alpha$ may have different effects across model families. A value that is large enough to affect one model may be too small to affect another. To make steering strength comparable across models (normalization), we calibrate $\alpha$ using the empirical scale of nonzero SAE activations. For model $m$, let $\bar z_m$ denote the mean nonzero SAE latent activation on reference prompts. We define the calibrated steering strength as
\[
\alpha_m = \frac{\alpha_{\mathrm{ref}}}{\bar z_{\mathrm{ref}}}\bar z_m.
\label{eq:alpha_norm}
\]
Here, \(\alpha_{\mathrm{ref}}\) and \(\bar z_{\mathrm{ref}}\) are measured on a reference model, so \(\alpha_m\) scales with the target model's SAE activation magnitude \footnote{Hyperparameter 2: We use an optimal steering value from GPT-2 as the base steering value ($\alpha_{ref} = 5$) as we have studied this model across a number of steering targets}.

\subsection{Contamination}
\label{sec:contamination}

\textsc{ContrastiveSteer} can still fail when features are weakly aligned, polysemantic, or too broadly activated. We therefore define contamination as a diagnostic that measures how much post-steering SAE activation is associated with weakly aligned features: specifically defined as $1-s_i$. The rationale is intuitive: target-aligned features contribute little to the score, while weakly aligned features contribute more. For a generated sequence of length $T$, with post-steering SAE activations $z'_{t,i}$, contamination is
\begin{equation}
\mathrm{Contamination}(y)
= \frac{1}{T\,d_{\mathrm{SAE}}}
\sum_{t=1}^{T}
\sum_{i=1}^{d_{\mathrm{SAE}}}
z'_{t,i}(1{-}s_i)
\label{eq:contamination}
\end{equation}

This definition gives the following interpretation: high contamination means that post-steering activation is not concentrated on high-relevance features, but instead remains weighted toward features with low target relevance. This can happen when the target topic is not cleanly represented in the SAE, when high-scoring features are still polysemantic, or when the steering is too weak to shift activation toward target-aligned features. Low contamination indicates that post-steering activation is more concentrated on features with high contrastive scores. Because contamination is based on the activation weights, values above 1 can occur when there is severe over-steering collapse (e.g. $\alpha$ = 2 for a model, layer, and topic where empirical $\alpha_{opt}=1$).

\section{Evaluation}

\begin{table*}[ht]
\centering
\setlength{\tabcolsep}{3pt}
\renewcommand{\arraystretch}{1.05}
\small
\begin{tabular}{l ccc ccc ccc}
\toprule
 & \multicolumn{3}{c}{\textbf{Medical}} & \multicolumn{3}{c}{\textbf{Financial}} & \multicolumn{3}{c}{\textbf{Legal}} \\
\cmidrule(lr){2-4}\cmidrule(lr){5-7}\cmidrule(lr){8-10}
\textbf{Method} & DS$\uparrow$ & Dist$\downarrow$ & NLI\,\%$\uparrow$ & DS$\uparrow$ & Dist$\downarrow$ & NLI\,\%$\uparrow$ & DS$\uparrow$ & Dist$\downarrow$ & NLI\,\%$\uparrow$ \\
\midrule
Orig.                & $.140{\pm}.003$ & $1.311{\pm}.002$ & $6$   & $.174{\pm}.005$ & $1.284{\pm}.004$ & $7$    & $.141{\pm}.005$ & $1.310{\pm}.004$ & $11$ \\
\textsc{Clamp}       & $.148{\pm}.003$ & $1.305{\pm}.003$ & $7$   & $.195{\pm}.011$ & $1.268{\pm}.008$ & $7$    & $.186{\pm}.004$ & $1.274{\pm}.003$ & $12$ \\
\textsc{P.Swap}      & $.140{\pm}.009$ & $1.311{\pm}.007$ & $9$   & $.175{\pm}.013$ & $1.284{\pm}.010$ & $17$   & $.182{\pm}.010$ & $1.278{\pm}.008$ & $14$ \\
\textbf{\textsc{CS}} & $\mathbf{.381{\pm}.006}$ & $\mathbf{1.112{\pm}.006}$ & $\mathbf{97}$ & $\mathbf{.514{\pm}.006}$ & $\mathbf{0.985{\pm}.006}$ & $\mathbf{100}$ & $\mathbf{.554{\pm}.008}$ & $\mathbf{.943{\pm}.008}$ & $\mathbf{42}$ \\
\bottomrule
\end{tabular}
\caption{Results using GPT-2 32k, best layer per domain, $\boldsymbol{\alpha}{=}5$, 5 random seeds (repetitions) across n=20 prompts,with 95\%, comparing \textsc{ContrastiveSteer} (\textsc{CS}) results with baselines including \textsc{P.Swap}:  \textsc{ProximitySwap}.}
\label{tab:contrast_main}
\label{tab:domain_clf}
\end{table*}

\begin{table*}[t]
\centering
\small
\setlength{\tabcolsep}{4pt}
\renewcommand{\arraystretch}{1.12}
\begin{tabularx}{0.98\textwidth}{p{0.11\textwidth} p{0.18\textwidth} X}
\toprule
\textbf{Domain} & \textbf{Steering Setting} & \textbf{Output} \\
\midrule

\multirow[t]{2}{*}{\raggedright Medical}
& \textsc{ContrastiveSteer}
& \emph{differ across \colorbox{lilac}{\textbf{sex differences}} ($p>0$)... \colorbox{lilac}{\textbf{treatment}} of \colorbox{lilac}{\textbf{diseases}}...} \\

& Original
& \emph{...in my car and my friend said he wanted my keys...} \\

\addlinespace[0.35em]
\cmidrule{1-3}

\multirow[t]{2}{*}{\raggedright Financial}
& \textsc{ContrastiveSteer}
& \emph{\colorbox{lilac}{\textbf{EUR}}\,1 \colorbox{lilac}{\textbf{\$10\,USD}} \colorbox{lilac}{\textbf{\$15\,USD}}... \colorbox{lilac}{\textbf{investment capital \$1{,}100\,USD}}} \\

& Original
& \emph{...school calendar... what about the kids in your class?} \\

\addlinespace[0.35em]
\cmidrule{1-3}

\multirow[t]{2}{*}{\raggedright Legal}
& \textsc{ContrastiveSteer}
& \emph{\colorbox{lilac}{\textbf{appeal}}... \colorbox{lilac}{\textbf{grounds}}... \colorbox{lilac}{\textbf{Court proceedings}}... \colorbox{lilac}{\textbf{Court of Appeal}}...} \\

& Original
& \emph{...anything but my own. They were in my car...} \\

\bottomrule
\end{tabularx}
\caption{
Representative GPT-2 32k continuations. \textsc{ContrastiveSteer} introduces domain vocabulary (in lilac), while baseline remain off-topic across all domains.
}
\label{tab:qualitative}
\end{table*}

\label{sec:eval}
We evaluate whether score-guided feature selection improves SAE-based topic steering and whether contamination diagnoses when steering fails with these experimental parameters: 

Domain datasets: medical~\citep{jin2019pubmedqa} (MIT License), financial~\citep{nickmuchi2023financial} (datasets are CC BY 3.0), and legal~\citep{chalkidis2022lexglue} (datasets are CC by 4.0) for the sections we use.

Model families: GPT-2 small with an MIT license and 117M parameters ~\citep{radford2019language}, Pythia-70M with an Apache 2.0 license and 70M parameters ~\citep{biderman2023pythia}, Gemma-2~2B~\citep{team2024gemma} with 2.6B parameters, and Gemma-3-1B~\citep{gemmateam2025gemma3technicalreport} with 1B parameters. Gemma models have their own Terms of Use on HuggingFace that must be signed before using.

Publicly released SAEs spanning three architectures:  (TopK --32k/138k parameters (GPT \cite{OAISAE}), ReLU -- 32k parameters (Pythia \cite{EleutherAIPythia70MDeduped}), and JumpReLU -- 16-65k parameters (Gemma \cite{lieberum2024gemma, NeuronpediaGemma})) under the MIT and CC BY 4.0 licenses. 

All are distributed and accessed via HuggingFace or SAELens (MIT License) \cite{bloom2024saelens} and fall within the intended use of all found model cards. Below, we describe setting up \textsc{ContrastiveSteer}, comparison baselines, and evaluation metrics.

\textbf{Reference  and alignment sets.} We construct $\mathcal{P}_{\mathrm{ref}}$ by sampling 200 prompts from WikiText-2-raw-v1~\citep{merity2016pointer} (CC-by-SA), filtered to 60--400 characters to exclude headers/fragments. For each domain, $\mathcal{P}_{\mathrm{align}}$ is a set of up to 200 long-answer passages filtered to 80--500 characters due to fragments and text artifacts. The \texttt{<endoftext>} token is excluded from all prompts to avoid artifacts in activation summaries.
  
\textbf{Evaluation Prompts: } Held-out generation prompts are drawn from the Empathetic Dialogues dataset~\citep{rashkin2019empathetic}, which provides domain-neutral conversational fragments. The same n=20 held-out prompts are used across all  domains.

\textbf{Baselines: } We introduce three comparable baselines by adapting methods from prior literature. First, we implemented feature clamping \citep{templeton2024scaling}, to be comparable in the contrastive scoring setting. Given contrastive scores $s$, \textsc{Clamp} amplifies only the top-$k$ highest-scoring features by multiplying their existing activations by $(1+\boldsymbol{\alpha})$:$ z'_{t,i} = z_{t,i}(1+\boldsymbol{\alpha}), \qquad i \in \mathrm{Top}_k(s).$
\textsc{Clamp} only modifies features already active at generation time\footnote{Hyperparameter 3 : We set $k = 1\% d_{\mathrm{SAE}}$, so that this method can scale with SAE size without being too large. In the reported results, this corresponds to 320 features for a 32k SAE.}.

Second, \textsc{ProximitySwap} uses proximity-based scoring as an alternative feature relevance signal. Each SAE feature is scored by its calibrated activation-weighted mean distance to $\mathcal{P}_{\mathrm{align}}$ across $\mathcal{P}_{\mathrm{ref}}$. Rather than adding to feature activations, \textsc{ProximitySwap} rewrites the active feature set while preserving original activation magnitudes:
$z'_t = \gamma(p_t) \odot \mathbb{I} \left[\sigma\!\left(\gamma(p_t) \odot s\right) \neq 0 \right]$,
where $s$ is the proximity score vector and $\odot$ denotes elementwise multiplication.
Third, the \textsc{Unsteered} baseline generates from the original model with the SAE, but without modifying SAE activations (no steering).

\textbf{Metrics:} We use the following metrics for evaluating outputs: domain score (DS\,$\uparrow$) is the mean maximum cosine similarity to $\mathcal{P}_{\text{align}}$,  topic distance (Dist\,$\downarrow$) is minimum Sentence-BERT (Apache 2.0 license) \cite{reimers-2019-sentence-bert} distance to $\mathcal{P}_{\text{align}}$,  contamination (Contam\,$\downarrow$) measures reliance on weakly aligned features, and Natural Language Inference Classification accuracy (NLI\,$\uparrow$) is zero-shot top-1 classification accuracy from \cite{he2021debertav3} (Apache 2.0 license), which we use within its intended use to measure whether outputs are classified as the target domain. Distinct-2, $d_2$, measures output language diversity as the fraction of unique bigrams in the generated text

\section{Results}
\subsection{ \textsc{ContrastiveSteer} Output Metrics}
\label{sec:crossmodel}

We compare \textsc{ContrastiveSteer} against the aforementioned baselines, domains, and model families at the layers that have been identified in the literature as being highly steerable (e.g. GPT-2 small \citep{radford2019language} with the OpenAI v5 32k TopK SAE~\citep{OAISAE} at layers~8--9 and $\boldsymbol{\alpha}{=}5$). Each experiment was repeated 5 times with n=20 held-out prompts per datasets, and 128 tokens generated as completions per prompt with top-$k$ sampling ($k{=}5$).

\begin{figure}[h]
\centering
\textbf{\textsc{ContrastiveSteer} $\Delta$DS at Calibrated $\boldsymbol{\alpha}$ Across Model Families}\\[4pt]
\includegraphics[width=\linewidth]{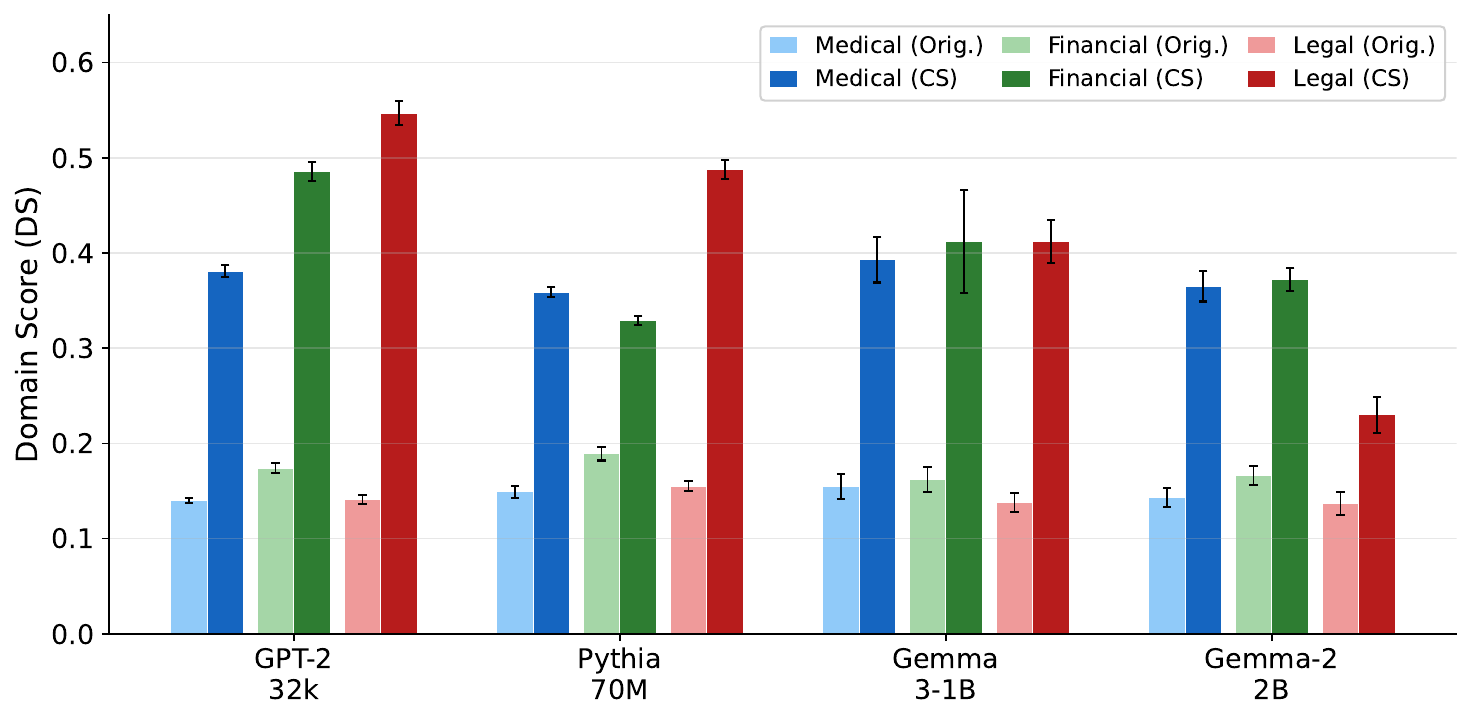}
\caption{\textsc{ContrastiveSteer} $\Delta$DS at calibrated $\boldsymbol{\alpha}$ across four model families and three domains. All models reach substantial gains over their unsteered baseline. }
\label{fig:crossmodel_bar}
\end{figure}

First, as shown in Fig.~\ref{fig:crossmodel_bar}, the change in domain score ($\Delta$DS) with the \textsc{ContrastiveSteer} calibrated $\alpha$ at the best steering layers per model is a considerable improvement over the unsteered \textsc{Original} baseline, ranging from 1.5x - over 3x on this metric. GPT-2 and Pythia-70M show results at the best layer per domain (GPT-2: L8/L7/L9; Pythia: L1/L2, Gemma-3-1B L17, Gemma-2-2b L13).

\textsc{ContrastiveSteer} also outperforms all baselines on all three metrics across domains, as shown in Tab.~\ref{tab:contrast_main} (on previous page) for GPT2 results. On embedding-based domain score and topic distance, gains are 2.7--3.9$\times$ over unsteered, which is supported by the NLI classification results where \textsc{ContrastiveSteer} outputs are classified as the target domain 42--100\% of the time, compared to 6--17\% for all baselines. This is a 4--16$\times$ improvement. In contrast, baseline methods like (\textsc{Clamp}, \textsc{ProximitySwap}) produce NLI scores indistinguishable from unsteered, showing that their modest DS gains likely do not reflect genuine topic steering.

\subsection{Steering Calibration Ablations}
\label{sec:alpha_calibration}

Our experiments support intuition from prior work that fixed steering strength does not transfer across model families: a value of $\alpha$ effective for GPT-2 or Pythia-70M can be entirely ineffective for Gemma-3-1B because the Gemma-3-1B SAE latent activations have a mean nonzero magnitude ${\approx}280$ (roughly $160{\times}$ larger than GPT-2/Pythia). Thus, steering $\alpha s_i$ at low $\alpha$, yields near-zero domain-relevant steering.

\begin{table}[ht]
\centering
\small
\setlength{\tabcolsep}{2pt}
\renewcommand{\arraystretch}{1.05}
\begin{tabular}{lcr>{\columncolor{gray!12}}r>{\columncolor{gray!12}}r}
\toprule
\textbf{Model / SAE} & mode $\alpha_{\mathrm{opt}} $ & $\alpha_m$ & \multicolumn{1}{c}{\shortstack{$\Delta$DS\\\scriptsize@$\alpha_{\mathrm{opt}}$}} & \multicolumn{1}{c}{\shortstack{$\Delta$DS\\\scriptsize@$\alpha_{m}$}} \\
\midrule
GPT-2 / 32k TopK    & 5   & $5.0$ & $+0.241$ & $+0.224$ \\
Pythia-70M / 32k    & 5   & $5.0$ & $+0.210$ & $+0.210$ \\
Gemma-3-1B / 65k    & 500  & $792$ & $+0.239$ & $+0.232$ \\
Gemma-2 2B / 16k    & 30  & $17.6$ & $+0.207$ & $+0.206$ \\
\bottomrule
\end{tabular}
\caption{
$\Delta$DS@$\alpha_m$ recovers 93-99\% of empirical grid-search $\Delta$DS@$\alpha_{\mathrm{opt}}$ across model families.
}
\label{tab:crossmodel}
\end{table}

\begin{table}[ht]
\centering
\small
\setlength{\tabcolsep}{4pt}
\renewcommand{\arraystretch}{1.08}
\begin{tabular}{c|ccc}
\toprule
\multicolumn{4}{l}{\textit{GPT-2 small} (12 layers, $\alpha{=}5$)} \\
\textbf{Layer} & \textbf{Medical} & \textbf{Financial} & \textbf{Legal} \\
\midrule
Orig & $.144{\pm}.002$ & $.171{\pm}.003$ & $.136{\pm}.001$ \\
 5   & $.331{\pm}.005$ & $.453{\pm}.007$ & $.404{\pm}.012$ \\
 7   & $.310{\pm}.003$ & $\mathbf{.514{\pm}.008}$ & $.541{\pm}.010$ \\
 8   & $\mathbf{.377{\pm}.017}$ & $.489{\pm}.005$ & $.497{\pm}.009$ \\
 9   & $.362{\pm}.007$ & $.481{\pm}.007$ & $\mathbf{.554{\pm}.010}$ \\
11   & $.144{\pm}.011$ & $.171{\pm}.016$ & $.136{\pm}.007$ \\
\midrule
\multicolumn{4}{l}{\textit{Pythia-70M} (6 layers, $\alpha{=}5$)} \\
\textbf{Layer} & \textbf{Medical} & \textbf{Financial} & \textbf{Legal} \\
\midrule
Orig &  $.149{\pm}.003$ & $.189{\pm}.003$ & $.155{\pm}.002$ \\
 1   & $.290{\pm}.003$ & $\mathbf{.329{\pm}.007}$ & $.379{\pm}.006$ \\
 2   & $\mathbf{.359{\pm}.008}$ & $.266{\pm}.004$ & $\mathbf{.488{\pm}.015}$ \\
 3   & $.355{\pm}.010$ & $.262{\pm}.005$ & $.464{\pm}.020$ \\
 4   & $.325{\pm}.007$ & $.296{\pm}.008$ & $.421{\pm}.010$ \\
 5   & $.149{\pm}.009$ & $.189{\pm}.011$ & $.155{\pm}.008$ \\
\bottomrule
\end{tabular}
\caption{DS (\textsc{ContrastiveSteer}) across layers}
\label{tab:layer_ablation}
\end{table}

\begin{table*}[t]
\centering
\small
\setlength{\tabcolsep}{5pt}
\renewcommand{\arraystretch}{1.05}
\begin{tabular}{llccccc}
\toprule
\textbf{Model / SAE} & \textbf{SAE type} & $\rho$ & $\boldsymbol{\alpha}_{\mathrm{opt}}$ & $\Delta$\textbf{Medical} & $\Delta$\textbf{Financial} & $\Delta$\textbf{Legal} \\
\midrule
\multicolumn{7}{l}{\textit{GPT-2, $\boldsymbol{\alpha}{=}5$, best layer, $n{=}5$ seeds --- SAE width ablation}} \\[2pt]
GPT-2 / 32k TopK    & TopK     & $0.719$ & 5   & $.241{\pm}.006$ & $.312{\pm}.010$ & $\mathbf{.406{\pm}.013}$ \\
GPT-2 / 128k TopK   & TopK     & $0.391$ & 5   & $.229{\pm}.010$ & $.190{\pm}.012$ & $.384{\pm}.007$ \\
\midrule
\multicolumn{7}{l}{\textit{Cross-family, each model at $\boldsymbol{\alpha}_{\mathrm{opt}}$, best layer}} \\[2pt]
Pythia-70M / 32k    & ReLU     & $0.26$--$0.36^\dagger$ & 5   & $.210{\pm}.005$ & $.140{\pm}.005$ & $.333{\pm}.010$ \\
Gemma-3-1B / 65k    & JumpReLU & $0.348$ & $500^\ddagger$ & $\mathbf{.238}$ & $\mathbf{.402}$ & $.328$ \\
Gemma-2 2B / 16k    & JumpReLU & $0.968$ & $30^\ddagger$ & $.207$ & $.228$ & $.091$ \\
\bottomrule
\end{tabular}
\caption{$\Delta$ Domain Score (DS) for \textsc{ContrastiveSteer} by SAE configuration and model family. Wider SAEs had lower coverage, but even still,  steered outputs exceed their unsteered counterparts.}
\label{tab:coverage_ablation}
\label{tab:width_ablation}
\label{tab:app_pythia}
\end{table*}

In this evaluation, we test if our proposed activation-scaling approach $\boldsymbol{\alpha}_m$ recovers most of the empirically identified optimum ($\boldsymbol{\alpha}_{opt}$) without any computationally-intensive generation-based search.Tab.~\ref{tab:crossmodel} compares $\boldsymbol{\alpha}_{\mathrm{opt}}$ with $\boldsymbol{\alpha}_m$ ($\alpha_{ref} = 5$ and ${\bar z_{\mathrm{ref}}} = 1.792$, calibrated using GPT-2). Using our method, $\boldsymbol{\alpha}_m$ recovers \textit{93--99\%} of $\Delta$DS@$\boldsymbol{\alpha}_{\mathrm{opt}}$ across all four model families.  This result suggests that much of the apparent cross-model instability in SAE steering comes from mismatched activation scales rather than from the feature-scoring rule alone. In practice, this means a new model/SAE pair can be calibrated with a reference forward pass, instead of requiring generation-time grid search over $\alpha$.

Beyond $\Delta$DS, the output text for models with the activation difference, like Gemma, was visible. For example, in the Gemma-3-1B financial setting, fixed-strength steering at $\alpha=5$ produces no visible topic steering: the steered continuation remains a generic narrative about ``driving to a funeral'' and ``meet her at a restaurant,''. After scale calibration at $\alpha_m=500$, the same method produces financial-domain language, including ``company,'' ``4-yr agreement,'' ``\$5.65B,'' ``\$4.7M net profit,'' and ``business.''

\begin{figure}[h]
\centering
\includegraphics[width=\columnwidth]{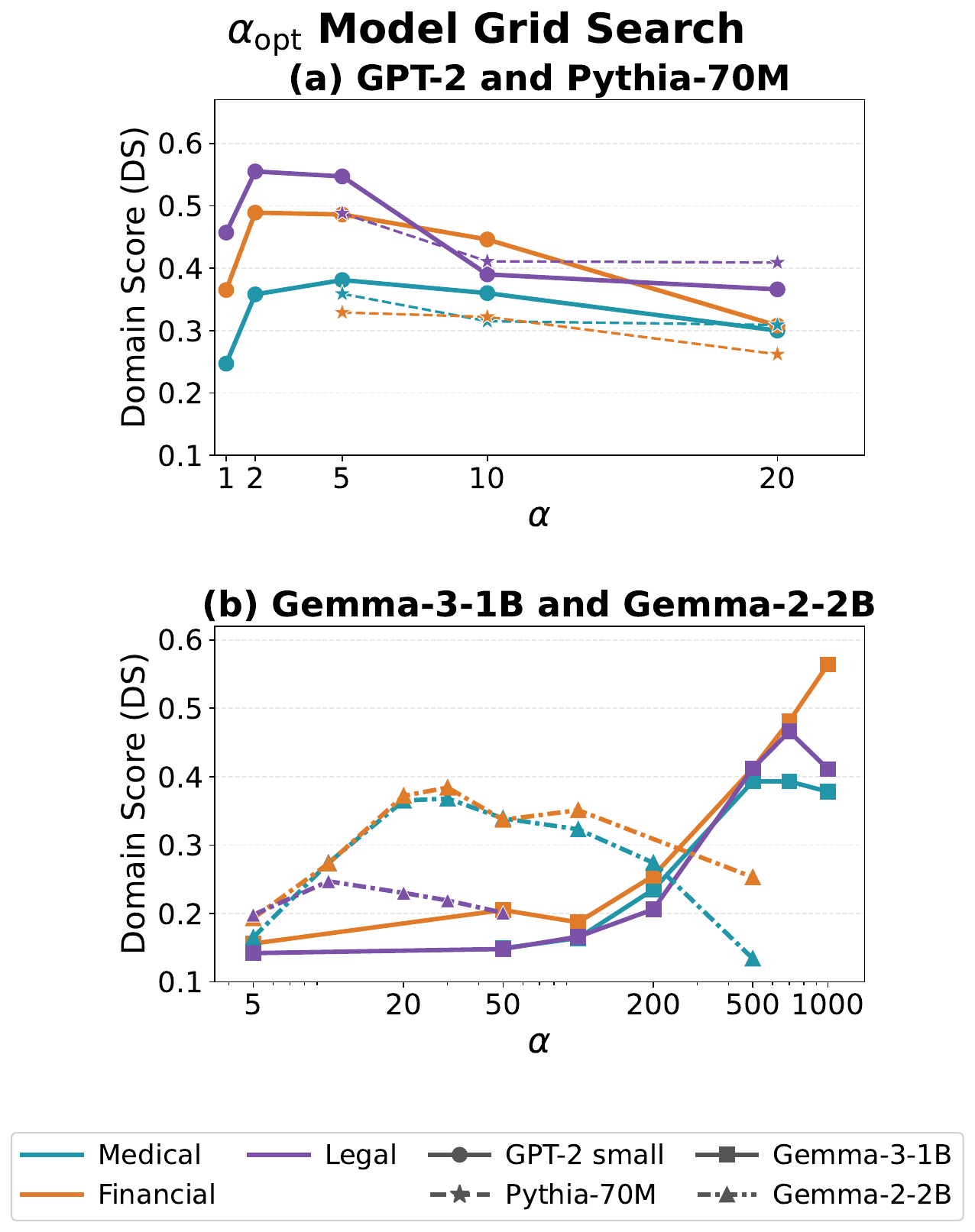}
\caption{
DS vs.\ $\alpha$ from per-model grid search. (a) GPT-2 and Pythia-70M at best layer per domain. (b) Gemma models on log scale, where available (otherwise steering collapse).
}
\label{fig:alpha_cal}
\end{figure}

Within a model, $\boldsymbol{\alpha}_{\mathrm{opt}}$ can vary but less dramatically, as shown in Fig.~\ref{fig:alpha_cal}. For different domains, on GPT-2 Fig.~\ref{fig:alpha_cal}a, performance is maximized in the $\alpha\in\{2,5\}$ range, while larger values reduce performance across domains. The Pythia-70M empirical grid search results shows a similar pattern: $\alpha{=}5$ is optimal at all 12 non-collapsed layer$\times$domain combination. These optimal values differ substantially from those needed for Gemma, shown in Fig.~\ref{fig:alpha_cal}b.

Still, as shown in Tab. \ref{tab:layer_ablation} , the value of $\alpha$ and overall steering change across layers. As shown in previous works \citep{gao2024scaling}, the final layers of a model are less steerable, and about 2/3rds of the way through the model is typically steerable. Models like Pythia with fewer layers may exhibit different properties, like that steering was more effective in the earlier layers.

\subsection{Contamination as Steering Diagnostic}

Finally, we evaluate Contamination (Ctm) (Eq.~\ref{eq:contamination}, Sec.~\ref{sec:contamination}), which measures whether post-steering activations remain associated with weakly aligned features,  across different SAE architectures and their feature coverage using $\mathcal{P}_{ref}$, also referred to as $\rho$. As Tab.~\ref{tab:coverage_ablation} shows \textsc{ContrastiveSteer}, within GPT-2, the wider 128k TopK SAE achieves lower coverage ($\rho{=}0.391$ vs.\ $0.719$) because TopK activation distributes sparsely over the larger feature space. Domain scores  also fall 5-39\% below the 32k baseline despite having $4{\times}$ more features. Across model families, coverage varies from 0.26--0.36 (Pythia/Gemma-3) to 0.97 (Gemma-2 2B), yet all families reach substantial $\Delta$DS once $\boldsymbol{\alpha}$ is calibrated, supporting that the SAE architecture itself, not just feature coverage, is important to calibrate over. Thus, feature coverage is useful for assessing whether scores can be estimated reliably, but \textit{not sufficient} to predict steering performance alone (without considering activation scale and SAE architecture).

\begin{table}[h]
\centering
\small
\setlength{\tabcolsep}{3pt}
\renewcommand{\arraystretch}{1.08}
\begin{tabularx}{\linewidth}{X c c c c p{0.18\linewidth}}
\toprule
\textbf{Setting} & $\boldsymbol{\alpha}$ & \textbf{Ctm} & $\boldsymbol{\Delta}$\textbf{DS} & $\boldsymbol{d_2}$ & \textbf{Regime} \\
\midrule
\rowcolor{yellow!22}
GPT-2, medical & 5 & .002 & +.237 & .376 & Success \\
\rowcolor{yellow!22}
GPT-2, financial & 5 & .002 & +.315 & .078 & Success; low $d_2$ \\
\rowcolor{yellow!22}
GPT-2, legal & 5 & .002 & +.413 & .189 & Success \\
\rowcolor{yellow!22}
Pythia-70M, medical & 5 & .001 & +.210 & .103 & Success \\
\midrule
\rowcolor{orange!20}
Gemma-3-1B, medical & 50 & .332 & $\approx$0 & .720 & Under-steered \\
\rowcolor{yellow!22}
Gemma-3-1B, medical & 500 & .301 & +.238 & .576 & Success \\
\midrule
\rowcolor{yellow!22}
Gemma-2 2B, medical & 30 & .270 & +.207 & .325 & Success \\
\rowcolor{red!14}
Gemma-2 2B, legal & 100 & 1.029 & $-.075$ & .230 & Over-steered \\
\bottomrule
\end{tabularx}
\caption{
Representative contamination regimes using Ctm and complementary heuristics, $\Delta$DS and $d_2$.
}
\label{tab:contam_diagnostic}
\end{table}

Tab.~\ref{tab:contam_diagnostic} shows that these metrics can divide the steered output into three categories. To do so, we use metrics $\Delta$DS, which measures whether the output becomes more target-aligned, and Distinct-2, $d_2$, measures output language diversity as the fraction of unique bigrams in the generated text to complement the contamination heuristic. First, at $\boldsymbol{\alpha}{\ll}\bar{z}_m$ (Gemma-3-1B, $\alpha{=}50$), contamination is high ($0.332$), $\Delta$DS is near zero, and $d_2$ remains high ($0.720$), indicating fluent but effectively unsteered output. Second, scaling to $\boldsymbol{\alpha}_{\mathrm{opt}}{=}500$ recovers large $\Delta$DS while preserving reasonably high language diversity ($d_2{=}0.576$), which suggests a topic shift in the output text rather than repetitive token injection, even though contamination remains higher than in GPT-2/Pythia. GPT-2 and Pythia achieve successful steering with comparable $\Delta$DS and near-zero contamination. Third is when $\boldsymbol{\alpha}$ is too high, like Gemma-2 2B legal at $\alpha{=}100$ metric values, which indicate over-steering collapse rather than useful domain shift.

\section{Conclusion}
\label{sec:conclusion}

\textsc{ContrastiveSteer} substantially improves topic alignment over the unsteered baseline across medical, financial, and legal domains. In addition to the contrastive-scoring based feature selection, this method required $\boldsymbol{\alpha}$ calibrated to the model's SAE latent activation scale --  the same method that fails on Gemma-3-1B at $\boldsymbol{\alpha}{=}5$ recovers large gains once $\boldsymbol{\alpha}$ is matched, and our proposed activation-scale normalization approach generalized across all evaluated model families without per-model tuning. Finally, we showed that contamination provides a lightweight diagnostic for when steering is ineffective and can be insightful in identifying successful steering, under-steering, and over-steering collapse  respectively.

\section*{Limitations}
\label{sec:discussion}

We made some assumptions in our experimental setup. Our experiments on model families and SAEs were constrained by publicly open source models and their respective SAEs. These SAEs are compute expensive to train \cite{gao2024scaling}, and there is not yet a consensus on the best approaches for training these SAEs for feature steering. Thus, whether calibration and contamination properties generalize to current frontier-scale models remains an open question. In addition, our metrics, like domain score, capture domain vocabulary and topic orientation, but not semantic coherence or factual accuracy. While the zero-shot NLI classifier provides an independent confirmation, it does not substitute for human evaluation, which should be the next step in addressing if this approach can be feasible for practitioners.

We also found  a fluency--alignment tradeoff worth exploring further, where steered topics do lose fluency when they are oversteered. \textsc{ContrastiveSteer} substantially increases perplexity at the settings that maximize domain score, but \textsc{ProximitySwap} and \textsc{Clamp} preserve fluency much better. This tradeoff is also worth investigating so that practitioners choosing between these methods can identify methods for their application. 

Finally, there are some risks associated with this method. Any topics can be steered, including potentially harmful and sensitive topics, which can also make this approach a vector for generating harmful text across multiple categories, although this is a prohibited use of our code.  Additional spot checks from the authors on the datasets were used to determine suitability of benchmarks and generated texts, although it is possible that the authors may have missed some generation texts, especially in $\mathcal{P}_{ref}$ and $\mathcal{P}_{align}$. In our search, we did not encounter any text that needed to be anonymized, especially within the medical domain. All datasets are in English, although generations can be in any character. 

\section*{Ethical Considerations}
 All experiments in main paper results use publicly available models and datasets. Code and supplementary materials will be released under the MIT License. AI assistants were used in the following categories: assistance with the language and grammar of the paper, porting code structure so that it was compatible with multiple tested models. No IRB review was conducted as this is not a human studies research project. 

\bibliography{sample-base}

@article{bayat2025steering,
  title={Steering large language model activations in sparse spaces},
  author={Bayat, Reza and Rahimi-Kalahroudi, Ali and Pezeshki, Mohammad and Chandar, Sarath and Vincent, Pascal},
  journal={arXiv preprint arXiv:2503.00177},
  year={2025}
}

@article{liu2023pre,
  title={Pre-train, Prompt, and Recommendation: A Comprehensive Survey of Language Modeling Paradigm Adaptations in Recommender Systems},
  author={Liu, Peng and Zhang, Lemei and Gulla, Jon Atle},
  journal={Transactions of the Association for Computational Linguistics},
  volume={11},
  pages={1553--1571},
  year={2023},
  publisher={MIT Press One Broadway, 12th Floor, Cambridge, Massachusetts 02142, USA~…}
}

@book{templeton2024scaling,
  title={Scaling monosemanticity: Extracting interpretable features from claude 3 sonnet},
  author={Templeton, Adly},
  year={2024},
  publisher={Anthropic},
  url={https://transformer-circuits.pub/2024/scaling-monosemanticity/}, 
  note={\url{https://transformer-circuits.pub/2024/scaling-monosemanticity/}}
}

@article{belinkov2021probingclassifierspromisesshortcomings,
  title={Probing classifiers: Promises, shortcomings, and advances},
  author={Belinkov, Yonatan},
  journal={Computational Linguistics},
  volume={48},
  number={1},
  pages={207--219},
  year={2022}
}

@inproceedings{gao2024scaling,
  title={Scaling and evaluating sparse autoencoders},
  author={Gao, Leo and Dupre la Tour, Tom and Tillman, Henk and Goh, Gabriel and Troll, Rajan and Radford, Alec and Sutskever, Ilya and Leike, Jan and Wu, Jeffrey},
  booktitle={International Conference on Learning Representations},
  volume={2025},
  pages={26721--26754},
  year={2025}
}

@article{turner2023activation,
  title={Activation addition: Steering language models without optimization},
  author={Turner, Alexander Matt and Thiergart, Lisa and Leech, Gavin and Udell, David and Mini, Ulisse and MacDiarmid, Monte},
  year={2024}
}

@article{zou2023representation,
 title={Representation Engineering: A Top-Down Approach to AI Transparency}, 
      author={Andy Zou and Long Phan and Sarah Chen and James Campbell and Phillip Guo and Richard Ren and Alexander Pan and Xuwang Yin and Mantas Mazeika and Ann-Kathrin Dombrowski and Shashwat Goel and Nathaniel Li and Michael J. Byun and Zifan Wang and Alex Mallen and Steven Basart and Sanmi Koyejo and Dawn Song and Matt Fredrikson and J. Zico Kolter and Dan Hendrycks},
      year={2025},
      eprint={2310.01405},
      archivePrefix={arXiv},
      primaryClass={cs.LG},
      url={https://arxiv.org/abs/2310.01405}, 
}

@inproceedings{arad2025saes,
  title={Saes are good for steering--if you select the right features},
  author={Arad, Dana and Mueller, Aaron and Belinkov, Yonatan},
  booktitle={Proceedings of the 2025 Conference on Empirical Methods in Natural Language Processing},
  pages={10252--10270},
  year={2025}
}

@article{chalnev2024improving,
  title={Improving steering vectors by targeting sparse autoencoder features},
  author={Chalnev, Sviatoslav and Siu, Matthew and Conmy, Arthur},
  journal={arXiv preprint arXiv:2411.02193},
  year={2024}
}

@article{team2024gemma,
  title={Gemma: Open Models Based on Gemini Research and Technology}, 
      author={Gemma Team and Thomas Mesnard and Cassidy Hardin and Robert Dadashi and Surya Bhupatiraju and Shreya Pathak and Laurent Sifre and Morgane Rivière and Mihir Sanjay Kale and Juliette Love and Pouya Tafti and Léonard Hussenot and Pier Giuseppe Sessa and Aakanksha Chowdhery and Adam Roberts and Aditya Barua and Alex Botev and Alex Castro-Ros and Ambrose Slone and Amélie Héliou and Andrea Tacchetti and Anna Bulanova and Antonia Paterson and Beth Tsai and Bobak Shahriari and Charline Le Lan and Christopher A. Choquette-Choo and Clément Crepy and Daniel Cer and Daphne Ippolito and David Reid and Elena Buchatskaya and Eric Ni and Eric Noland and Geng Yan and George Tucker and George-Christian Muraru and Grigory Rozhdestvenskiy and Henryk Michalewski and Ian Tenney and Ivan Grishchenko and Jacob Austin and James Keeling and Jane Labanowski and Jean-Baptiste Lespiau and Jeff Stanway and Jenny Brennan and Jeremy Chen and Johan Ferret and Justin Chiu and Justin Mao-Jones and Katherine Lee and Kathy Yu and Katie Millican and Lars Lowe Sjoesund and Lisa Lee and Lucas Dixon and Machel Reid and Maciej Mikuła and Mateo Wirth and Michael Sharman and Nikolai Chinaev and Nithum Thain and Olivier Bachem and Oscar Chang and Oscar Wahltinez and Paige Bailey and Paul Michel and Petko Yotov and Rahma Chaabouni and Ramona Comanescu and Reena Jana and Rohan Anil and Ross McIlroy and Ruibo Liu and Ryan Mullins and Samuel L Smith and Sebastian Borgeaud and Sertan Girgin and Sholto Douglas and Shree Pandya and Siamak Shakeri and Soham De and Ted Klimenko and Tom Hennigan and Vlad Feinberg and Wojciech Stokowiec and Yu-hui Chen and Zafarali Ahmed and Zhitao Gong and Tris Warkentin and Ludovic Peran and Minh Giang and Clément Farabet and Oriol Vinyals and Jeff Dean and Koray Kavukcuoglu and Demis Hassabis and Zoubin Ghahramani and Douglas Eck and Joelle Barral and Fernando Pereira and Eli Collins and Armand Joulin and Noah Fiedel and Evan Senter and Alek Andreev and Kathleen Kenealy},
      year={2024},
      eprint={2403.08295},
      archivePrefix={arXiv},
      primaryClass={cs.CL},
      url={https://arxiv.org/abs/2403.08295}, 
  note={\url{https://huggingface.co/google/gemma-2b} accessed: 2025-05-25}
}

@inproceedings{lieberum2024gemma,
  title={Gemma scope: Open sparse autoencoders everywhere all at once on gemma 2},
  author={Lieberum, Tom and Rajamanoharan, Senthooran and Conmy, Arthur and Smith, Lewis and Sonnerat, Nicolas and Varma, Vikrant and Kram{\'a}r, J{\'a}nos and Dragan, Anca and Shah, Rohin and Nanda, Neel},
  booktitle={Proceedings of the 7th BlackboxNLP Workshop: Analyzing and Interpreting Neural Networks for NLP},
  pages={278--300},
  year={2024}, 
  note={\url{https://www.neuronpedia.org/gemma-scope#steer} Accessed 2026-03-20}
}

@article{sanh2021multitask,
  title={Multitask Prompted Training Enables Zero-Shot Task Generalization}, 
      author={Victor Sanh and Albert Webson and Colin Raffel and Stephen H. Bach and Lintang Sutawika and Zaid Alyafeai and Antoine Chaffin and Arnaud Stiegler and Teven Le Scao and Arun Raja and Manan Dey and M Saiful Bari and Canwen Xu and Urmish Thakker and Shanya Sharma Sharma and Eliza Szczechla and Taewoon Kim and Gunjan Chhablani and Nihal Nayak and Debajyoti Datta and Jonathan Chang and Mike Tian-Jian Jiang and Han Wang and Matteo Manica and Sheng Shen and Zheng Xin Yong and Harshit Pandey and Rachel Bawden and Thomas Wang and Trishala Neeraj and Jos Rozen and Abheesht Sharma and Andrea Santilli and Thibault Fevry and Jason Alan Fries and Ryan Teehan and Tali Bers and Stella Biderman and Leo Gao and Thomas Wolf and Alexander M. Rush},
      year={2022},
      eprint={2110.08207},
      archivePrefix={arXiv},
      primaryClass={cs.LG},
      url={https://arxiv.org/abs/2110.08207}, 
  note={ Accessed via \url{https://huggingface.co/datasets/bigscience/P3} on 2026-05-25}
}

@inproceedings{reimers-2019-sentence-bert,
  title = "Sentence-BERT: Sentence Embeddings using Siamese BERT-Networks",
  author = "Reimers, Nils and Gurevych, Iryna",
  booktitle = "Proceedings of the 2019 Conference on Empirical Methods in Natural Language Processing",
  month = "11",
  year = "2019",
  publisher = "Association for Computational Linguistics",
  url = "https://arxiv.org/abs/1908.10084",
  note={Model used: \url{https://huggingface.co/sentence-transformers/all-MiniLM-L6-v2} Accessed: 2025-05-25}
}

@inproceedings{marks2024sparse,
  title={Sparse feature circuits: Discovering and editing interpretable causal graphs in language models},
  author={Marks, Samuel and Rager, Can and Michaud, Eric and Belinkov, Yonatan and Bau, David and Mueller, Aaron},
  booktitle={International Conference on Learning Representations},
  volume={2025},
  pages={23888--23923},
  year={2025}
}

@misc{OAISAE,
 author={OpenAI},
  title = {SAEViewer},
  howpublished = {\url{https://openaipublic.blob.core.windows.net/sparse-autoencoder/sae-viewer/index.html#/model/gpt2-small/family/v5_32k/layer/6/location/resid_post_mlp/feature/25975}},
  note = {Accessed: 2026-03-01}
}

@misc{NeuronpediaGemma,
author={Neuronpedia},
  title = {Exploring Gemma2 With Gemma Scope},
  howpublished = {\url{https://www.neuronpedia.org/gemma-2-2b/0-gemmascope-att-16k/101}},
  note = {Accessed: 2026-03-01}
}

@inproceedings{jin2019pubmedqa,
  title={Pubmedqa: A dataset for biomedical research question answering},
  author={Jin, Qiao and Dhingra, Bhuwan and Liu, Zhengping and Cohen, William and Lu, Xinghua},
  booktitle={Proceedings of the 2019 conference on empirical methods in natural language processing and the 9th international joint conference on natural language processing (EMNLP-IJCNLP)},
  pages={2567--2577},
  year={2019},
  note={\url{https://huggingface.co/datasets/qiaojin/PubMedQA} Accessed 2026-03-25}
}

@article{bereska2024mechanistic,
  title={Mechanistic Interpretability for AI Safety--A Review},
  author={Bereska, Leonard and Gavves, Efstratios},
  journal={arXiv preprint arXiv:2404.14082},
  year={2024}
}

@article{peng2025use,
  title={Use sparse autoencoders to discover unknown concepts, not to act on known concepts},
  author={Peng, Kenny and Movva, Rajiv and Kleinberg, Jon and Pierson, Emma and Garg, Nikhil},
  journal={arXiv preprint arXiv:2506.23845},
  year={2025}
}

@article{wu2025axbench,
  title={Axbench: Steering llms? even simple baselines outperform sparse autoencoders},
  author={Wu, Zhengxuan and Arora, Aryaman and Geiger, Atticus and Wang, Zheng and Huang, Jing and Jurafsky, Dan and Manning, Christopher D and Potts, Christopher},
  journal={arXiv preprint arXiv:2501.17148},
  year={2025}
}

@misc{guardrailoutputs,
  title={How to implement LLM guardrails},
  author={Jarvis, Colin},
  organization={OpenAI Cookbook},
  howpublished = "\url{https://cookbook.openai.com/examples/how_to_use_guardrails}",
  year={2023},
  month={Dec},
  day={19}
}

@misc{bloom2024saelens,
  title        = {{SAELens}},
  author       = {Bloom, Joseph and Tigges, Curt and Duong, Anthony and Chanin, David},
  year         = {2024},
  howpublished = {\url{https://github.com/decoderesearch/SAELens}},
  note         = {Accessed: 2026-05-25}
}

@misc{EleutherAIPythia70MDeduped,
  author       = {{EleutherAI}},
  title        = {{Pythia-70M-Deduped}},
  year         = {2023},
  howpublished = {\url{https://www.neuronpedia.org/pythia-70m-deduped/5-res-sm}},
  note         = {Accessed: 2026-05-25}
}

@misc{gemmateam2025gemma3technicalreport,
      title={Gemma 3 Technical Report}, 
      author={Gemma Team and Aishwarya Kamath and Johan Ferret and Shreya Pathak and Nino Vieillard and Ramona Merhej and Sarah Perrin and Tatiana Matejovicova and Alexandre Ramé and Morgane Rivière and Louis Rouillard and Thomas Mesnard and Geoffrey Cideron and Jean-bastien Grill and Sabela Ramos and Edouard Yvinec and Michelle Casbon and Etienne Pot and Ivo Penchev and Gaël Liu and Francesco Visin and Kathleen Kenealy and Lucas Beyer and Xiaohai Zhai and Anton Tsitsulin and Robert Busa-Fekete and Alex Feng and Noveen Sachdeva and Benjamin Coleman and Yi Gao and Basil Mustafa and Iain Barr and Emilio Parisotto and David Tian and Matan Eyal and Colin Cherry and Jan-Thorsten Peter and Danila Sinopalnikov and Surya Bhupatiraju and Rishabh Agarwal and Mehran Kazemi and Dan Malkin and Ravin Kumar and David Vilar and Idan Brusilovsky and Jiaming Luo and Andreas Steiner and Abe Friesen and Abhanshu Sharma and Abheesht Sharma and Adi Mayrav Gilady and Adrian Goedeckemeyer and Alaa Saade and Alex Feng and Alexander Kolesnikov and Alexei Bendebury and Alvin Abdagic and Amit Vadi and András György and André Susano Pinto and Anil Das and Ankur Bapna and Antoine Miech and Antoine Yang and Antonia Paterson and Ashish Shenoy and Ayan Chakrabarti and Bilal Piot and Bo Wu and Bobak Shahriari and Bryce Petrini and Charlie Chen and Charline Le Lan and Christopher A. Choquette-Choo and CJ Carey and Cormac Brick and Daniel Deutsch and Danielle Eisenbud and Dee Cattle and Derek Cheng and Dimitris Paparas and Divyashree Shivakumar Sreepathihalli and Doug Reid and Dustin Tran and Dustin Zelle and Eric Noland and Erwin Huizenga and Eugene Kharitonov and Frederick Liu and Gagik Amirkhanyan and Glenn Cameron and Hadi Hashemi and Hanna Klimczak-Plucińska and Harman Singh and Harsh Mehta and Harshal Tushar Lehri and Hussein Hazimeh and Ian Ballantyne and Idan Szpektor and Ivan Nardini and Jean Pouget-Abadie and Jetha Chan and Joe Stanton and John Wieting and Jonathan Lai and Jordi Orbay and Joseph Fernandez and Josh Newlan and Ju-yeong Ji and Jyotinder Singh and Kat Black and Kathy Yu and Kevin Hui and Kiran Vodrahalli and Klaus Greff and Linhai Qiu and Marcella Valentine and Marina Coelho and Marvin Ritter and Matt Hoffman and Matthew Watson and Mayank Chaturvedi and Michael Moynihan and Min Ma and Nabila Babar and Natasha Noy and Nathan Byrd and Nick Roy and Nikola Momchev and Nilay Chauhan and Noveen Sachdeva and Oskar Bunyan and Pankil Botarda and Paul Caron and Paul Kishan Rubenstein and Phil Culliton and Philipp Schmid and Pier Giuseppe Sessa and Pingmei Xu and Piotr Stanczyk and Pouya Tafti and Rakesh Shivanna and Renjie Wu and Renke Pan and Reza Rokni and Rob Willoughby and Rohith Vallu and Ryan Mullins and Sammy Jerome and Sara Smoot and Sertan Girgin and Shariq Iqbal and Shashir Reddy and Shruti Sheth and Siim Põder and Sijal Bhatnagar and Sindhu Raghuram Panyam and Sivan Eiger and Susan Zhang and Tianqi Liu and Trevor Yacovone and Tyler Liechty and Uday Kalra and Utku Evci and Vedant Misra and Vincent Roseberry and Vlad Feinberg and Vlad Kolesnikov and Woohyun Han and Woosuk Kwon and Xi Chen and Yinlam Chow and Yuvein Zhu and Zichuan Wei and Zoltan Egyed and Victor Cotruta and Minh Giang and Phoebe Kirk and Anand Rao and Kat Black and Nabila Babar and Jessica Lo and Erica Moreira and Luiz Gustavo Martins and Omar Sanseviero and Lucas Gonzalez and Zach Gleicher and Tris Warkentin and Vahab Mirrokni and Evan Senter and Eli Collins and Joelle Barral and Zoubin Ghahramani and Raia Hadsell and Yossi Matias and D. Sculley and Slav Petrov and Noah Fiedel and Noam Shazeer and Oriol Vinyals and Jeff Dean and Demis Hassabis and Koray Kavukcuoglu and Clement Farabet and Elena Buchatskaya and Jean-Baptiste Alayrac and Rohan Anil and Dmitry and Lepikhin and Sebastian Borgeaud and Olivier Bachem and Armand Joulin and Alek Andreev and Cassidy Hardin and Robert Dadashi and Léonard Hussenot},
      year={2025},
      eprint={2503.19786},
      archivePrefix={arXiv},
      primaryClass={cs.CL},
      url={https://arxiv.org/abs/2503.19786}, 
      note={\url{https://huggingface.co/google/gemma-3-1b-pt}, accessed 2026-05-25}
}

@inproceedings{rashkin2019empathetic,
  title={Towards empathetic open-domain conversation models: A new benchmark and dataset},
  author={Rashkin, Hannah and Smith, Eric Michael and Li, Margaret and Boureau, Y-Lan},
  booktitle={Proceedings of the 57th annual meeting of the association for computational linguistics},
  pages={5370--5381},
  year={2019},
  note={\url{https://huggingface.co/papers/1811.00207}. Accessed: 2026-05-25}
}

@article{merity2016pointer,
  title={Pointer sentinel mixture models},
  author={Merity, Stephen and Xiong, Caiming and Bradbury, James and Socher, Richard},
  journal={arXiv preprint arXiv:1609.07843},
  year={2016},
  note={\url{https://huggingface.co/datasets/Salesforce/wikitext}, Accessed 2026-03-25}
}

@article{he2021debertav3,
  title={Debertav3: Improving deberta using electra-style pre-training with gradient-disentangled embedding sharing},
  author={He, Pengcheng and Gao, Jianfeng and Chen, Weizhu},
  journal={arXiv preprint arXiv:2111.09543},
  year={2021},
  note={The specific model used is: \url{https://huggingface.co/cross-encoder/nli-deberta-v3-small}. Accessed 2026-03-25}
}

@misc{nickmuchi2023financial,
  title={Financial Classification Dataset},
  author={Nick Muchi},
  year={2023},
  howpublished={\url{https://huggingface.co/datasets/nickmuchi/financial-classification}}
}

@article{radford2019language,
  title={Language models are unsupervised multitask learners},
  author={Radford, Alec and Wu, Jeffrey and Child, Rewon and Luan, David and Amodei, Dario and Sutskever, Ilya},
  journal={OpenAI blog},
  volume={1},
  number={8},
  pages={9},
  year={2019}
}

@inproceedings{biderman2023pythia,
  title={Pythia: A Suite for Analyzing Large Language Models Across Training and Scaling}, 
      author={Stella Biderman and Hailey Schoelkopf and Quentin Anthony and Herbie Bradley and Kyle O'Brien and Eric Hallahan and Mohammad Aflah Khan and Shivanshu Purohit and USVSN Sai Prashanth and Edward Raff and Aviya Skowron and Lintang Sutawika and Oskar van der Wal},
      year={2023},
      eprint={2304.01373},
      archivePrefix={arXiv},
      primaryClass={cs.CL},
      url={https://arxiv.org/abs/2304.01373}, 
  note={Used pythia-70m-deduped-res-sm from SAELens. Accessed 2026-03-25}
}

@inproceedings{chalkidis2022lexglue,
  title={LexGLUE: A benchmark dataset for legal language understanding in English},
  author={Chalkidis, Ilias and Jana, Abhik and Hartung, Dirk and Bommarito, Michael and Androutsopoulos, Ion and Katz, Daniel and Aletras, Nikolaos},
  booktitle={Proceedings of the 60th Annual Meeting of the Association for Computational Linguistics (Volume 1: Long Papers)},
  pages={4310--4330},
  year={2022},
  url={https://huggingface.co/datasets/coastalcph/lex_glue},
  notes={This is a conglomeration of benchmarks, accessed through HuggingFace and SAELens, which corresponds to the ECtHR dataset with authors: Ilias Chalkidis, Ion Androutsopoulos, and Nikolaos Aletras}
}

\newpage

\section{Appendix}

\begin{figure}[t]
    \centering
    \begin{subfigure}[t]{0.47\textwidth}
        \centering
        \includegraphics[width=0.8\textwidth]{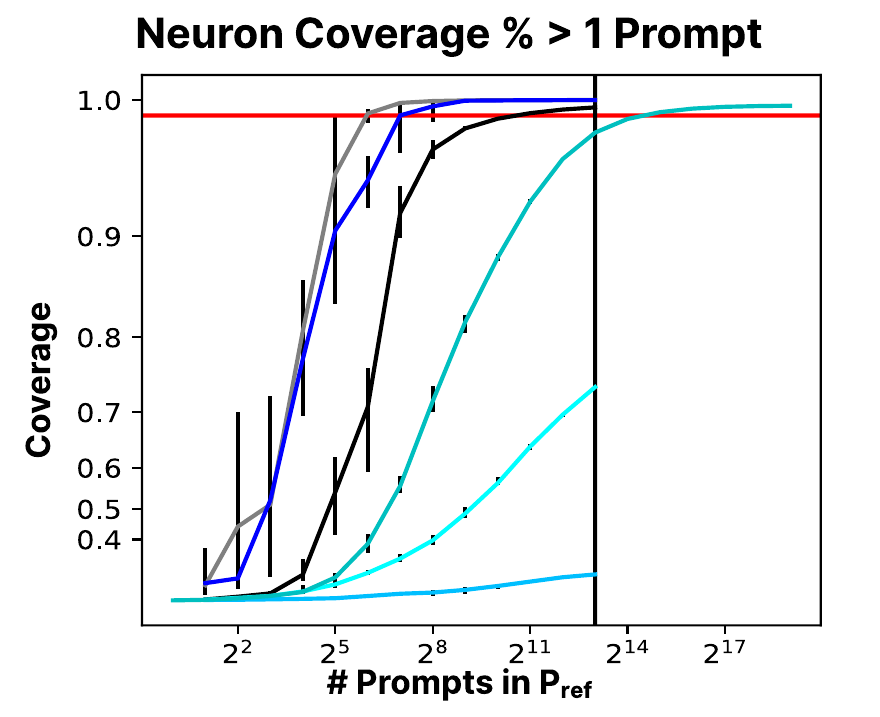}
        \caption{Fraction of features activated at least once.}
        \label{fig:subfig1}
    \end{subfigure}
    \hfill
    \begin{subfigure}[t]{0.47\textwidth}
        \centering
        \includegraphics[width=\textwidth]{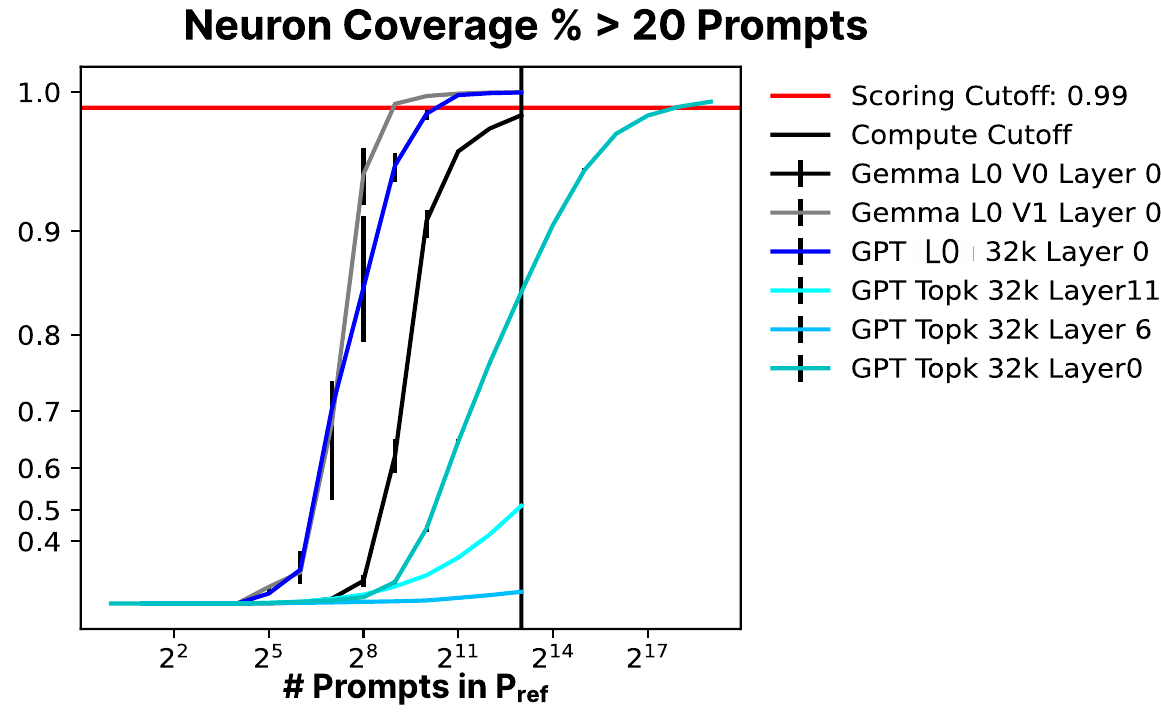}
        \caption{Fraction activated by ${\geq}20$ distinct prompts.}
        \label{fig:subfig22}
    \end{subfigure}
  \caption{SAE feature coverage as a function of reference set size $|\mathcal{P}_{\text{ref}}|$, for five SAE configurations spanning two models and multiple layers. Error bars are max/min over 5 random draws of each size. The right panel shows the subset used for score estimation (our 20-prompt threshold); GPT-2 32k and Gemma 65k differ substantially at the same corpus size.}
\label{fig4}
\end{figure}

\textbf{Reproducibility.} \textsc{ContrastiveSteering} code is provided in the supplementary zip file, implemented in PyTorch. Experiments were conducted using NVIDIA V100 Tesla GPUs, but across multiple timeframes. As a result, we estimate the total computation-related costs: several hundred GB for storage for experimentation in total, across 100-200 GPU hours. We used at most 2 parallel devices due to resource constraints. All code and supplementary materials are released under the MIT License and contains additional details about the models used, copyright, and terms of service.

\subsection{Feature Score and Layer-Analysis}
\label{app:score_validation}

Fig.~\ref{scores} and the following sections shows scoring distributions using \textsc{ProximitySteering} across SAE configurations and alignment set sizes. Larger $|\mathcal{P}_{\text{align}}|$ shifts scores upward, because more alignment prompts reduce the minimum distance from reference to alignment text, increasing the fraction of features with positive scores. Fig.~\ref{scoring2} shows that top-$k$ scoring features activate substantially more on $\mathcal{P}_{\text{align}}$ prompts than on unaligned text across models and layers, showing that high-scoring features are empirically associated with the target topic.

\paragraph{Additional $\mathcal{P}_{ref}$ and $\mathcal{P}_{align}$ experiments}
For four additional relatively distinct tasks from the P3 set from HuggingFace (Apache-2.0)~\citep{sanh2021multitask} (Amazon, Yelp, IMDB, and Wikipedia), we sampled 20 prompts and calculated inter- and intra-topic Euclidean distances, shown in Tab.~\ref{dist-table}. Intra-topic distance exceeds inter-topic distance for all four tasks, showing that Euclidean distance in Sentence-BERT \cite{reimers-2019-sentence-bert} embedding space separates topically similar from dissimilar prompts. Some cross-topic pairs fall within the intra-topic range, but these correspond to prompts sharing surface vocabulary despite different task labels.

\begin{table}[ht]
\centering
\begin{tabular}{lll}
{\bf Task}  &\multicolumn{2}{c}{\bf Distances}\\
{}&{Intra-topic}&{Inter-topic}\\
 \hline \\
Yelp & $1.357 \pm 0.005$ & $1.128 \pm 0.04$ \\
Wiki & $1.412 \pm 0.004$ & $1.144 \pm 0.041$ \\
Amazon & $1.34 \pm 0.006$ & $0.95 \pm 0.04$ \\
IMDB & $1.34 \pm 0.005$ & $1.00 \pm 0.04$ \\
\end{tabular}
\caption{Intra- vs.\ inter-topic Euclidean distances in Sentence-BERT embedding space for four P3 tasks, validating the choice of distance metric for proximity-based scoring.}
\label{dist-table}
\end{table}

\subsection{SAE width and coverage trade-off.}

To characterize how well different SAE configurations are covered by $\mathcal{P}_{\text{ref}}$, we sample five reference sets of increasing size (with replacement) and measure (i) the fraction of SAE features activated at least once, and (ii) the fraction of features activated by more than 20 distinct prompts.

Since open-source SAE releases are recent \citep{gao2024scaling, lieberum2024gemma}, there is limited prior characterization of how consistently different SAE configurations activate across realistic text distributions, which directly affects whether feature-topic scores can be estimated reliably. As shown in Fig.~\ref{fig4}, coverage varies substantially across SAE configurations, including differences in the underlying model and layer, activation function $\sigma$, and SAE width.
In particular, several layer-0 SAEs achieve near-complete activation coverage (above 0.99) with approximately $2^{14} \approx 8$K prompts, while deeper-layer SAEs exhibit lower coverage at comparable sample sizes, consistent with the presence of rare or effectively dead features \citep{gao2024scaling}.

Because the alignment scores estimate whether a feature preferentially activates on prompts near $\mathcal{P}_{\text{align}}$, features that activate too rarely do not provide enough evidence to distinguish aligned behavior from polysemanticity or noise.
Accordingly, we exclude SAE features for which fewer than 20 reference prompts yield nonzero activation, and compute topic-relevance scores only over the remaining subset.

\subsection{Intermediate Layer-Level Results}
\label{app:layer_results}

An ablation for \textsc{ProximitySwap} multiplies feature alignment scores by token activations without applying the masking step; this leads to high reconstruction error instability, suggesting that unmasked score-weighted activation is a poor modification mechanism.

\begin{figure}[h]
    \centering
    \begin{subfigure}[t]{0.4\textwidth}
        \centering
        \includegraphics[width=\textwidth,height=5.5cm,keepaspectratio=false]{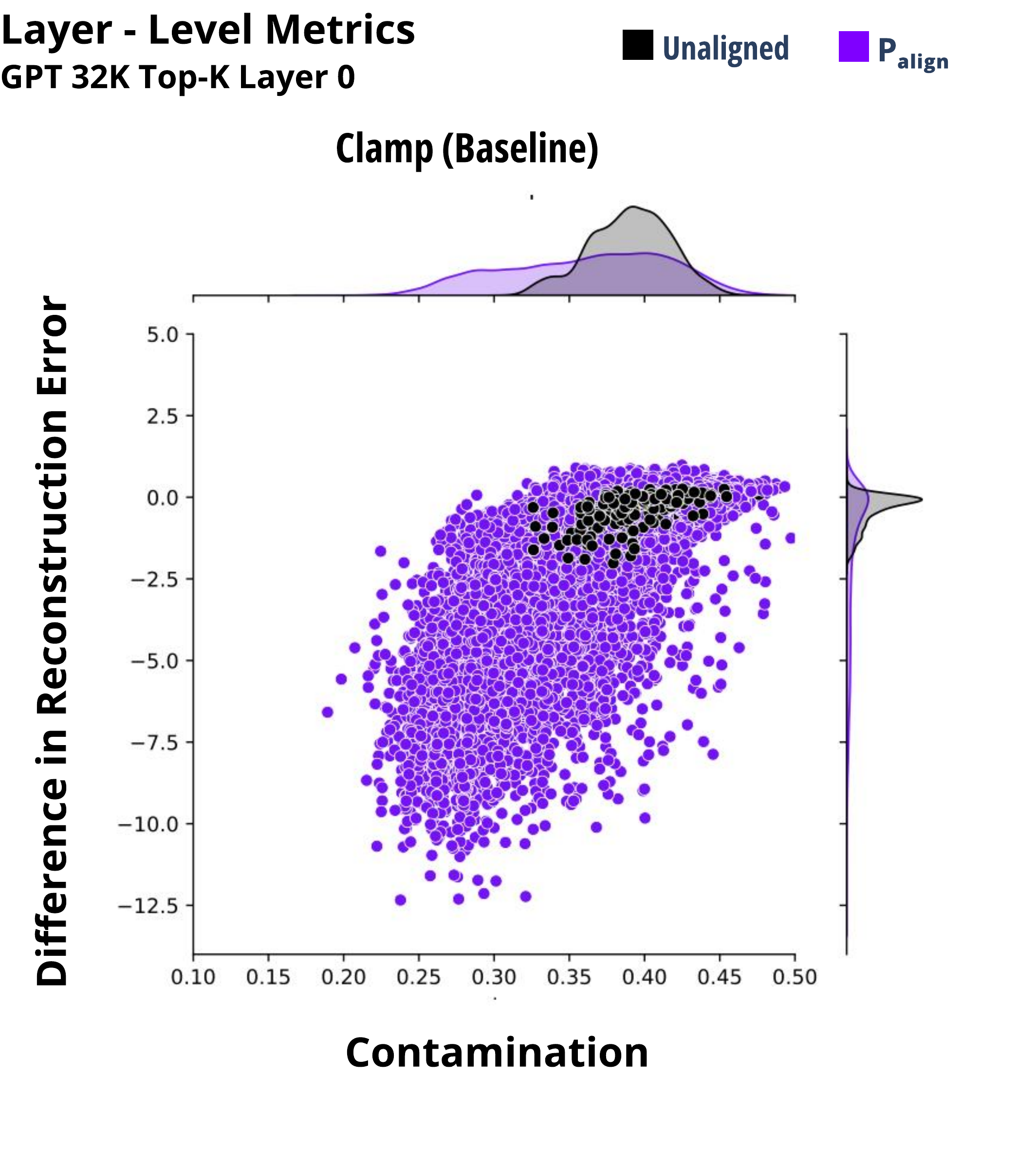}
        \caption{Metrics for \textsc{Clamp} approach.}
        \label{fig:subfig4c}
    \end{subfigure}
    \hfill
    \begin{subfigure}[t]{0.4\textwidth}
        \centering
        \includegraphics[width=\textwidth,height=5.5cm,keepaspectratio=false]{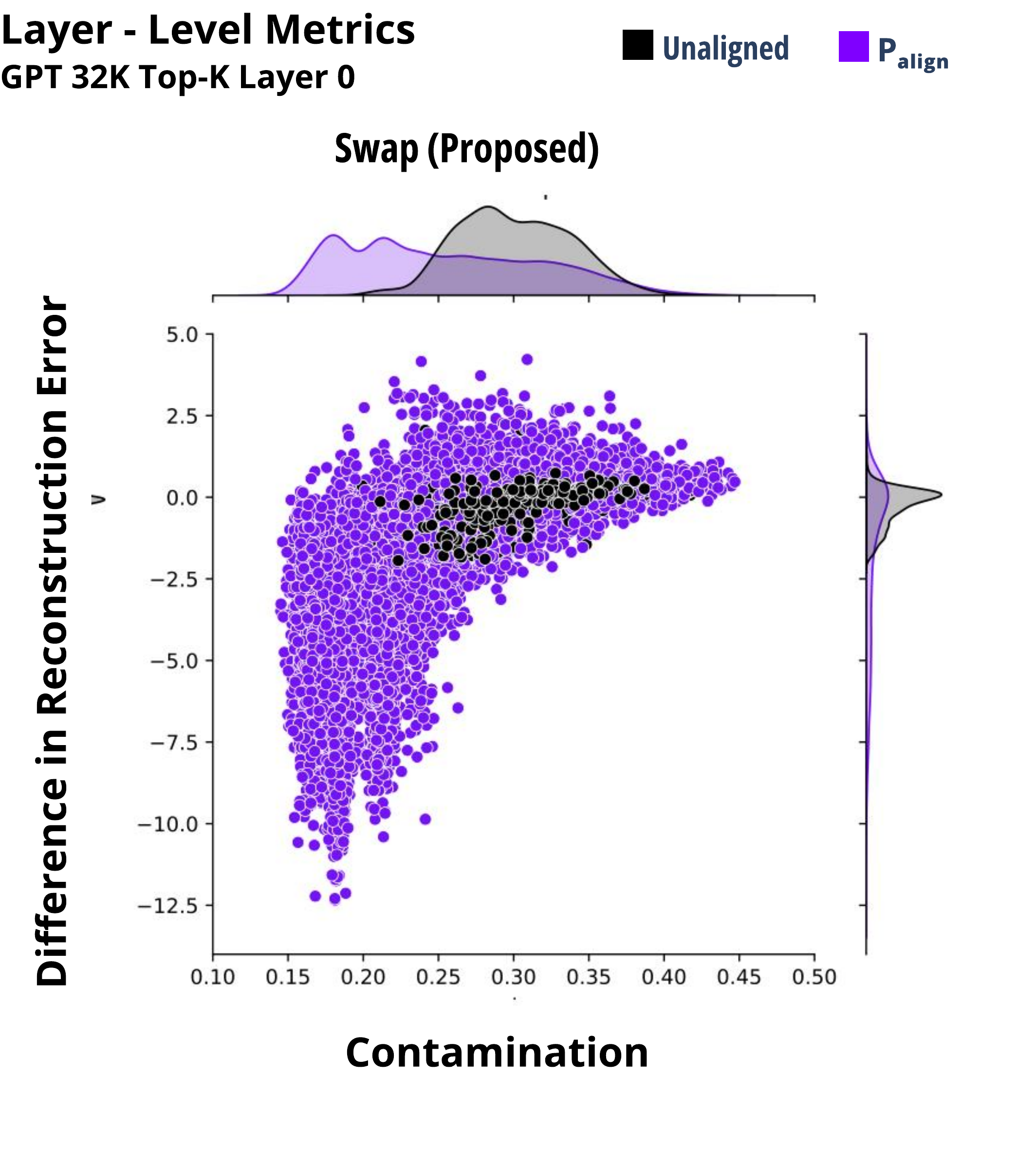}
        \caption{\textsc{ProximitySwap}: reconstruction error difference and contamination per token.}
        \label{fig:subfig3s}
    \end{subfigure}
\caption{Reconstruction error difference (relative to unmodified SAE output) and contamination per token for a variation of \textsc{Clamp} \textbf{(a)} and \textsc{ProximitySwap} \textbf{(b)}. Aligned tokens show smaller reconstruction error changes; \textsc{ProximitySwap} achieves lower contamination than \textsc{Clamp}. Special tokens not shown.}
\label{pptimg}
\end{figure}

\begin{figure*}[]
\centering
\includegraphics[width=\textwidth]{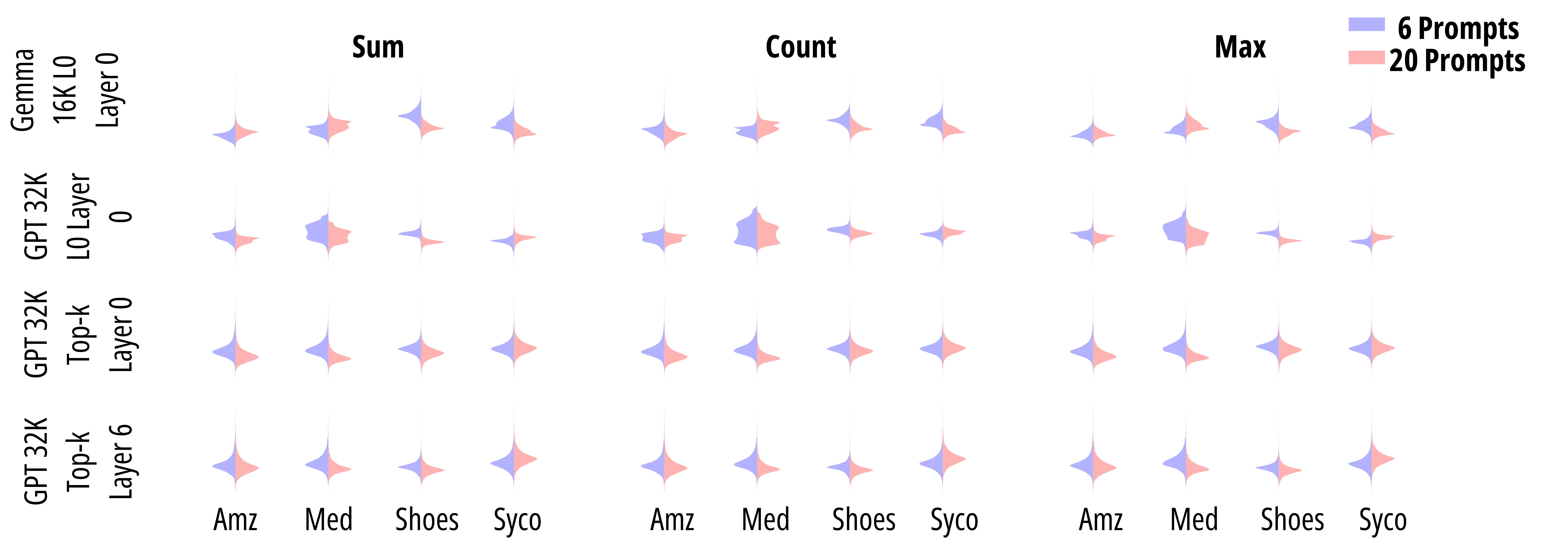}
\caption{Distributions of SAE feature scores (Eq.~\ref{eq:cscore}) across SAE configurations (model, layer, activation function) and alignment set sizes ($|\mathcal{P}_{\text{align}}|{=}20$--100 prompts). Larger alignment sets shift scores upward, reflecting smaller minimum distances between reference and alignment prompts.}
\label{scores}
\end{figure*}

To evaluate the effect of these modifications on layer outputs, we consider two heuristics: (1) the difference in reconstruction error relative to the original SAE output, and (2) the contamination measure introduced earlier. Fig.~\ref{pptimg} shows that aligned tokens generally exhibit lower differences in reconstruction error than unaligned tokens, indicating that modifications preserve token-level structure more effectively in aligned contexts. Unaligned tokens exhibit higher reconstruction error differences, which is expected given that the SAE is trained to minimize reconstruction error over general text rather than alignment-specific modifications.

Across SAE configurations, a variation \textsc{ProximitySwap} consistently exhibits lower contamination than \textsc{Clamp}, indicating that feature selection under \textsc{ProximitySwap} relies more heavily on features with higher empirical alignment scores. While both approaches introduce deviations from the original SAE output, these heuristics suggest that context-sensitive feature selection reduces contributions from weakly aligned features.

\subsection{Additional Computational Cost Tables}
\label{app:compute}

Tab.~\ref{infcost} breaks down SAE setup cost for these classes of general steering methods on the NVIDIA hardware. Feature scoring has lower per-topic setup cost than fine-tuning (62\,s vs.\ 333.6\,s, no gradient computation required), though current SAE inference is slower per generated token (6.20--6.64\,s/token vs.\ 0.40\,s/token for fine-tuning) due to the SAE forward pass.

\begin{table}[h]
\centering
\small
\setlength{\tabcolsep}{4pt}
\renewcommand{\arraystretch}{1.1}
\begin{tabular}{lccc}
\toprule
\textbf{Task} & \textbf{s/tok} & \textbf{Mode} & \textbf{Phase} \\
\midrule
Init Embed.   & $0.07{\pm}0.001$ & Parallel & \multirow{2}{*}{Setup} \\
Ref. Gen.     & $0.02{\pm}0.001$ & Parallel & \\
\midrule
\rowcolor{gray!15} Align Embed.  & $0.07{\pm}0.001$ & Parallel & \multirow{3}{*}{\cellcolor{gray!15}Per-topic} \\
\rowcolor{gray!15} Dist. Calc    & $3{\times}10^{-4}$ & Sequential & \\
\rowcolor{gray!15} Feat. Scoring & $4{\times}10^{-4}$ & Sequential & \\
\bottomrule
\end{tabular}
\caption{Compute breakdown: setup tasks (white) run once per corpus; per-topic tasks (gray) run once per target domain.}
\label{infcost}
\end{table}

\subsection{Layer-Level $\alpha_m$ Calibration}
\label{app:layer_alpha}

Tab.~\ref{tab:alpha_layer_compare} shows $\Delta$DS at $\boldsymbol{\alpha}_m$ per layer and domain, showing that the calibrated value generalises across layers, not just across model families.

\begin{table}[!ht]
\centering
\small
\setlength{\tabcolsep}{4pt}
\renewcommand{\arraystretch}{1.08}
\begin{tabular}{lc|ccc|c}
\toprule
\textbf{Model} & \textbf{Lyr} & \textbf{Med} & \textbf{Fin} & \textbf{Leg} & \textbf{Match?} \\
\midrule
\multicolumn{6}{l}{\textit{Pythia-70M, $\boldsymbol{\alpha}_m{=}5$}} \\[1pt]
& L1 & $+.141$ & $+.140$ & $+.224$ & \checkmark \\
& L2 & $+.210$ & $+.077$ & $+.333$ & \checkmark \\
& L3 & $+.206$ & $+.073$ & $+.309$ & \checkmark \\
& L4 & $+.176$ & $+.107$ & $+.266$ & \checkmark \\
& L5 & ${\approx}0$ & ${\approx}0$ & ${\approx}0$ & collapse \\
\midrule
\multicolumn{6}{l}{\textit{GPT-2, $\boldsymbol{\alpha}_m{=}5$, representative layers}} \\[1pt]
& L7 & $+.170$ & $+.340$ & $+.400$ & \checkmark \\
& L8 & $+.237$ & $+.315$ & $+.356^\dagger$ & $\dagger$ (86\%) \\
& L9 & $+.222$ & $+.307$ & $+.413$ & \checkmark \\
\bottomrule
\end{tabular}
\caption{$\Delta$DS at $\boldsymbol{\alpha}_m$ (activation-scale-normalised) per layer and domain.}
\label{tab:alpha_layer_compare}
\end{table}

\begin{figure*}[]
\centering
\includegraphics[width=\textwidth]{Frame_6-4.pdf}
\caption{Activation levels on top-$k$ scoring features for $\mathcal{P}_{\text{align}}$ prompts (purple) vs.\ unaligned prompts (black), across SAE configurations. Separation between curves indicates topic relevance of the selected features.}
\label{scoring2}
\end{figure*}

\end{document}